\documentclass{article}
\usepackage{ijcai21}

\usepackage{times}
\usepackage{soul}
\usepackage{url}
\usepackage[hidelinks]{hyperref}
\usepackage[utf8]{inputenc}
\usepackage[small]{caption}
\usepackage{graphicx}
\usepackage{amsmath}
\usepackage{amsthm}
\usepackage{booktabs}
\usepackage{algorithm}
\usepackage{algorithmic}
\usepackage{multirow} %add by peijun
\usepackage{subcaption} %add by peijun
\usepackage{amsmath} %add by peijun
\usepackage{amssymb} %add by peijun
\usepackage{adjustbox} %add by peijun
\usepackage{threeparttable} %add by peijun 
\usepackage{comment} %add by peijun 

\title{Learning Sample Importance for Cross-Scenario Video Temporal Grounding}
\author{
	Peijun Bao,
	Yadong Mu\\
	\affiliations{
    Peking University, \\
    \{peijunbao, myd\}@pku.edu.cn
    }
}

\begin{document}

\maketitle

\begin{abstract}
The task of temporal grounding aims to locate video moment in an untrimmed video, with a given sentence query. 
This paper for the first time investigates some superficial biases that are specific to the temporal grounding task, and proposes a novel targeted solution. Most alarmingly, we observe that existing temporal ground models heavily rely on some biases (\emph{e.g.}, high preference on frequent concepts or certain temporal intervals) in the visual modal. This leads to inferior performance when generalizing the model in cross-scenario test setting. To this end, we propose a novel method called Debiased Temporal Language Localizer (Debias-TLL) to prevent the model from naively memorizing the biases and enforce it to ground the query sentence based on true inter-modal relationship. Debias-TLL simultaneously trains two models. By our design, a large discrepancy of these two models' predictions when judging a sample reveals higher probability of being a biased sample. Harnessing the informative discrepancy, we devise a data re-weighing scheme for mitigating the data biases. We evaluate the proposed model in cross-scenario temporal grounding, where the train / test data are heterogeneously sourced. Experiments show large-margin superiority of the proposed method in comparison with state-of-the-art competitors.
\end{abstract}

\section{Introduction}
Given a sentence query and an untrimmed video, the goal of temporal grounding~\cite{mcn,tall} is to localize video moment described by the sentence query. In recent years, a list of promising models~\cite{cbp,excl,wacv,rl,tempo,mm20,lifeifei,comp,scdm,2dmap} have been designed to tackle this task.
Despite remarkable research progress, we empirically find that these models are heavily affected by some superficial bias of the data, leading to inferior generalization performance on cross-scenario testing data. In one of our pilot experiments, we 
take some well-trained state-of-the-art temporal grounding model and zero the feature vector of all testing queries. This boils down to using only the visual information in the temporal grounding. Surprisingly, the performance under such a setting is comparable to many models that normally read both queries and videos during testing. It also significantly outstrips random guess.

\begin{figure}[t!]
\begin{center}
\includegraphics[width=0.48\textwidth]{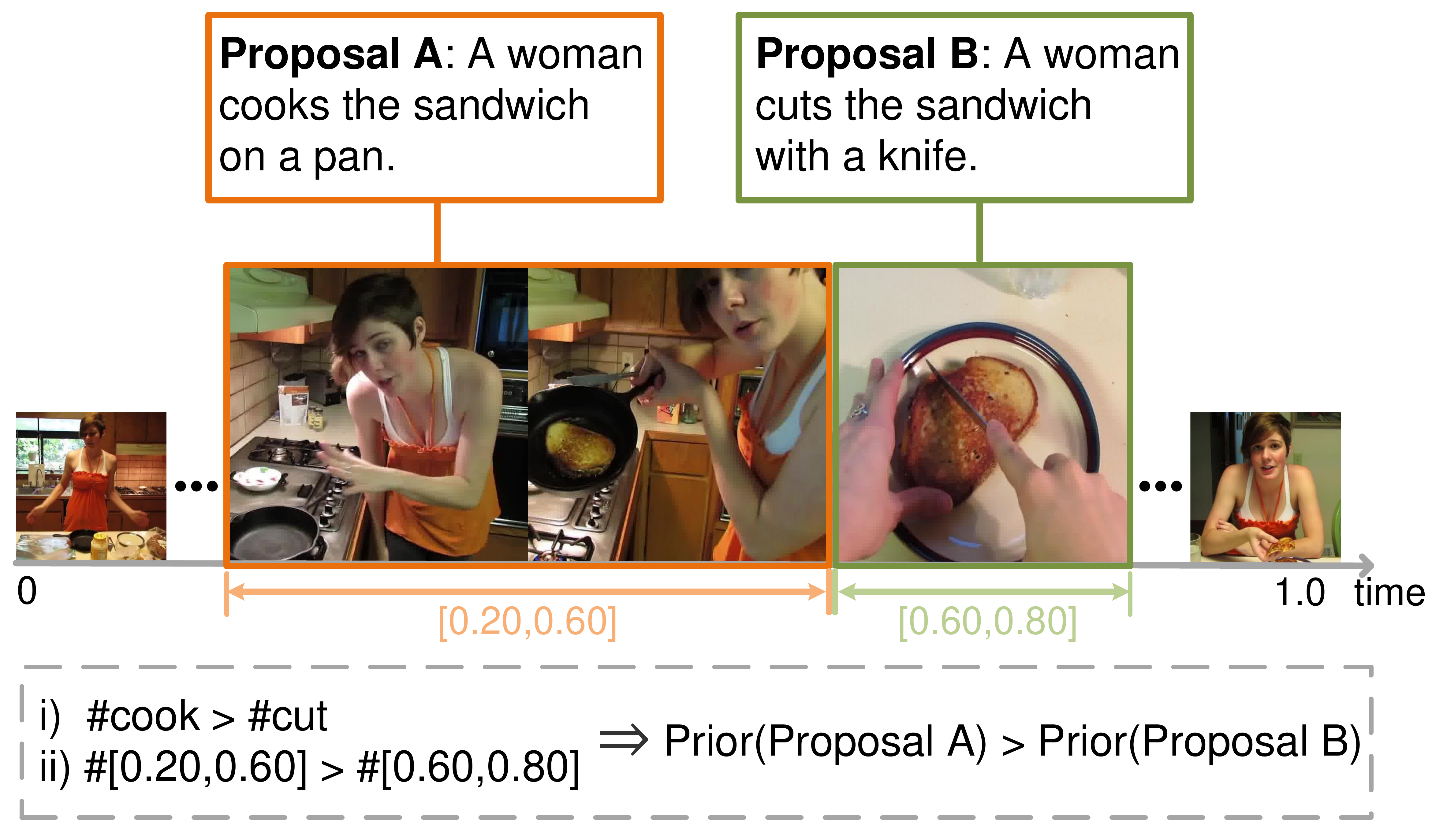}
\end{center}
\caption{\small Illustration of video uni-modal bias in ActivityNet. Even without knowing the sentence query, the prior of video moment proposal A to be grounded is larger than B, mainly due to two facts in the annotations: i) the visual concept ``cook'' contained in A  are much more frequently queried than the concept ``cut'' in B; ii) the temporal interval $[0.2,0.6]$ of A appears more frequently annotated than B. \# denotes the frequency.}\label{fig:motivation}
\vspace{-0.1in}
\end{figure}

To make the point more clear, let us provide some concrete empirical observations. We have identified two sorts of dominant biases that a model can exploit for over-fitting the training scenarios, namely the visual content bias and temporal interval bias. In detail, a few visual concepts and temporal intervals are  more frequently queried by the sentence than others in the training dataset. 
Although the issue of dominant biases have been previously reported as language bias in visual-question answering tasks~\cite{vqa_regularization,vqa_unimodal,vqa_counterfact}, the preference of visual contents and temporal intervals is specific to the temporal grounding task and our report here is the first. 
For instance, the concept ``run" is largely frequent than ``sit" in the benchmark of ActivityNet. Likewise, certain temporal intervals are more likely to be grounded. As illustrated in Figure~\ref{fig:motivation}, the interval $[0.20,0.60]$ statistically has more annotations than $[0.60,0.80]$. If above biases were sufficiently strong, a fully uni-modal input (such as zeroing query's features) can still achieve good performance in this multi-modal task. However, when generalizing the learned model into other unseen scenarios, these superficial biases between video moments and ground-truth may disappear, which adversely impacts the cross-scenario performance.

\begin{figure}[t!]
\begin{center}
\includegraphics[width=0.48\textwidth]{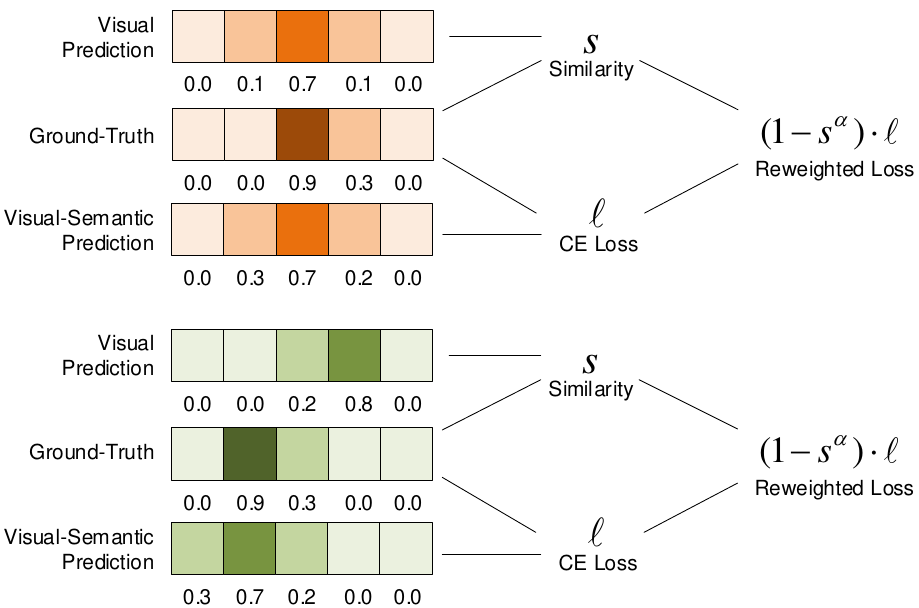}
\end{center}
\caption{\small Sample importance reweighing. 
The visual localizer adjusts the loss function for the visual-semantic localizer, \emph{i.e.}, suppressing the importance of training sample with high relevance to video moment bias (upper figure) and augmenting the weight of the irrelevant one (lower figure).
}\label{fig:key_idea}
\vspace{-0.2cm}
\end{figure}

To this end, we propose a novel method called Debiased Temporal Language Localizer (Debias-TLL) to prevent the model from naively learning the video moment bias and enforce it to ground sentence in the video. Our key idea is to simultaneously train two twined models, with one of them aiming to learn video moment bias from the data and further to debias the other model. The models have an identical backbone. One of them reads only the video input, and the other normally has access to the full video-query input. The first model is designed to learn the video moment bias and predict the localization results only from visual modality. As illustrated in Figure~\ref{fig:key_idea}, we then use the prediction of the first model to reweigh the importance of training samples for the second model and adjust the loss function accordingly. During this process, those training samples with high probability to being biased is suppressed. In this way, the training data is adaptively re-weighed in order to mitigate video moment bias in the second model. At the inference stage, we drop the first model and only use the second one for final prediction.

Note that the weakness of video modality biased model cannot be reflected by existing standard evaluation process because the training and testing data share a similar distribution of video moment correlation. To fairly evaluate the model, we propose a novel cross-scenario setting for the video temporal grounding task. In specific, we conduct the training and evaluation processes across two data distributions where video moment correlation cannot transfer from one data distribution to another. Under such settings, a model which makes prediction utilizing video moment bias would fail to perform well on the testing data.

Our contributions are summarized as follows:

1) To the best of our knowledge, we are the first to investigate the video moment bias in the video temporal grounding task, which adversely affects the generalization ability of the model. Two specific sorts of video moment biases in the data (visual content bias and temporal interval bias) are studied.

2) We propose a novel two-model-based methods to re-weigh the training data via learning sample importance and delineate the video moment bias for a temporal grounding model.

3) A cross-scenario evaluation setting is proposed to reveal the weakness of video-moment biased temporal grounding. Our proposed method beats the state-of-the-art competitors with a clear margin under the cross-scenario settings.

%%%%%%%%%%%%%%%%%%%%%%%%%%%%% Related Work %%%%%%%%%%%%%%%%%%%%%%%%%%%%%
\section{Related Work}
\subsection{Temporal Grounding}
The task of temporal grounding in video is recently introduced by~\cite{mcn,tall}, which aims to determine the start and end time of the video moment described by a sentence query. ~\cite{mcn} proposes a moment context network to jointly model text query and video clips. ~\cite{tall} proposes cross-modal localizer to regress action boundary for candidate video clips. ~\cite{acrn,attention_2} advice to apply attention mechanism to highlight the crucial part of visual features or query contents. ~\cite{sm_rl} then develops a semantic matching reinforcement learning framework to reduce the large visual-semantic discrepancy between video and language.

Several recent works~\cite{man,2dmap,cbp} propose to model temporal dependencies within sentence to closely integrate language and video representation via graph convolution or non-local modules. And ~\cite{mm20,comp,cmin} further utilize compositional property of query sentence and decompose sentence as multiple components for better temporal reasoning.

\begin{figure*}[t!]
\begin{center}
\includegraphics[width=0.95\textwidth]{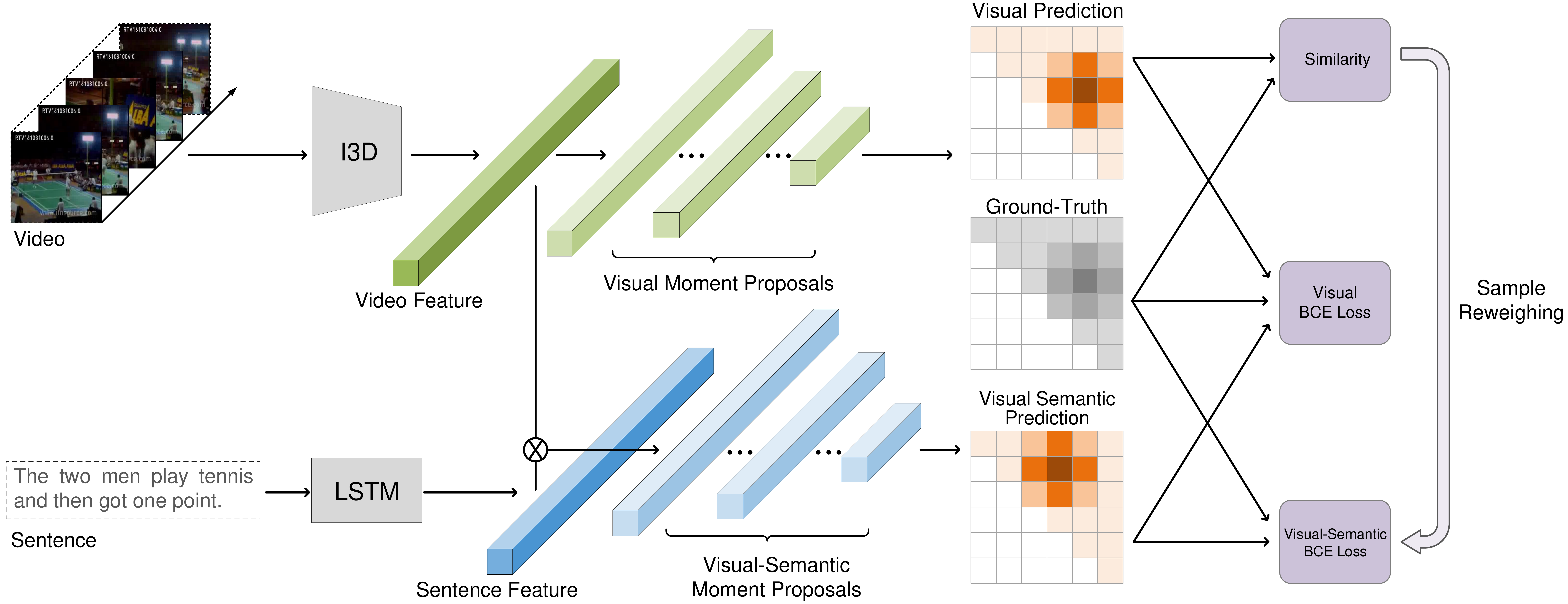}
\end{center}
%\vspace{-0.5cm}
\caption{\small Our proposed network consists of four main components: a language encoder, a video encoder, a visual localizer, and a visual-semantic localizer. The visual localizer learns the video moment bias from the data with a single input of video modality. Then it adjusts the loss function for the visual-semantic localizer, \emph{i.e.}, suppressing the importance of training sample with high relevance to video moment biases.
}\label{fig:methods}
%\vspace{-0.3cm}
\end{figure*}

\subsection{Unbiased Cross-Modal Understanding}
\cite{vqa_regularization,vqa_unimodal,vqa_counterfact} study language bias in visual question answering (VQA) caused by  answer prior. 
\cite{vqa_counterfact} proposes a model-agnostic counterfactual samples-synthesizing training scheme to reduce the language biases, which generates numerous counterfactual training samples by masking critical objects in images or words in questions.
\cite{vqa_regularization} introduces a question-only model, and then pose training as an adversarial game which discourages the VQA model from capturing language biases in its question.

Recently, \cite{unbias_scene_graph} studies the predicate bias on the task of scene graph generation from image.
And~\cite{visual_dialog} investigates the dialog history bias in the visual dialog  and proposes two causal principles for improving the quality of visual dialog.
To the best of our knowledge, the video moment bias is specific to the temporal grounding task and is never explored before.

\section{Methods}

\subsection{Problem Formulation}

Given an untrimmed video $V$ and a sentence description $S$, the goal of temporal grounding is to localize temporal moments $T$ described by the sentence. More specifically, the video is presented as a sequence of frames $V=\{v_{i}\}_{i=1}^{L_V}$ where $v_i$ is the feature of $i$-th frame and $L_V$ is the frame number of the video. The sentence description $S$ is presented as $S=\{s_{i}\}_{i=1}^{L_S}$ where $s_{i}$ represents $i$-th word in the sentence and $L_S$ denotes the total number of words. The temporal moment $T$ is defined by the start and end time points of the moment in the video.

\subsection{Debiased Temporal Language Localizer}
As illustrated in Figure~\ref{fig:methods}, our proposed Network consists of four main components: a language encoder, a video encoder, a visual-semantic localizer, and a visual localizer. This section will elaborate on the details of each component.

\subsubsection{Video Encoder}
Video encoder aims to obtain  high-level visual representations of video moment proposals from raw input frames.
Specifically, the input video is first segmented into small clips, with each video clip containing $T$ frames. A fixed-interval sampling is performed  to obtain $N$ video clips. For each sampled video clip, we extract a sequence of ordinary spatio-temporal features $V=\{v_{i}\}_{i=1}^{N}$  with a pretrained I3D Network~\cite{i3d}.

The visual feature embeddings for moment proposals are constructed from these basic I3D features. For a moment proposal ($a,b$) with start point at $a$ and end point at $b$,  we apply boundary-matching ($\mathrm{BM}$) operation ~\cite{bmn} over all I3D features covered by this proposal to get the feature embedding:
\begin{equation}\label{eq:proposal_f}
\tilde{f}^{V_{ab}}=\mathrm{BM} ( \{v_{i}\}_{i=a}^{b} ).
\end{equation}

The boundary-matching operation can efficiently generate proposal-level feature from basic clip-level feature, through a series of bilinear sampling and convolutional operations.
More algorithmic details are omitted here and can be referred to~\cite{bmn}. $\tilde{f}^{V_{ab}}$ is passed through a fully-connected layer to obtain the final feature embedding $f^{V_{ab}} \in \mathbb{R}^{d^V}$ for the moment proposal ($a,b$).
Essentially, this extracted feature $f^{V_{ab}}$ summarizes spatial-temporal patterns from raw input frame and thus represents the visual
structure of the moment proposal.

\subsubsection{Language Encoder}
Given an input of a natural language sentence query, the goal of language encoder is to encode the sentence such that moments of interest can be effectively retrieved  in the video.
Our language encoder extracts  feature embedding $f^{S}$ of the sentence descriptions $S$.

Instead of encoding each word with a one-hot vector or learning word embeddings from scratch, we rely on word embeddings obtained from a large collection of text documents.
In more details, each word $s_{i}$ in $S$ is first encoded into Glove word embedding~\cite{glove} as $w_{i}$.
Then the sequence of word embedding $\{w_{i}\}_{i=1}^{L_S}$ is fed to an LSTM~\cite{lstm}. The last hidden state of LSTM is passed to a single fully-connected layer to extract the final sentence feature $f^{S}\in \mathbb{R}^{d^S}$.

\subsubsection{Visual Localizer and Visual-Semantic Localizer}
We design two twined models i.e. visual localizer and visual-semantic localizer, 
with the visual localizer aiming to learn video moment bias from the data 
and further to debias the visual-semantic localizer.
The visual localizer reads only the video input, and the visual-semantic localizer normally has access to the full video-query input.

%In more details, 
The visual-semantic localizer first constructs visual-semantic features of moment proposals for the sentence query and then localizes the described moments. In specific, video moment feature  $f^{V_{ab}}$ for all possible moment proposals are computed according to Eq.~(\ref{eq:proposal_f}) where $1\leq a \leq b \leq N$.
The features of the visual modality and language modality are fused to generate visual-semantic features for each moment proposals. To interact the language feature $f^{S}$ with video moment feature  $f^{V_{ab}}$, 
we multiply $f^{S}$ with video moment clip feature $f^{V_{ab}}$ and then normalize the fused feature $\hat{M}_{ab}$ with its $\mathcal{L}_2$ norm, namely
\begin{equation}\label{eq:fuse}
    \begin{split}
	    & \hat{M}_{ab} = f^{V_{ab}} \odot f^{S},\\
	    & M_{ab} = \hat{M}_{ab}/||\hat{M}_{ab}||_2,\\
	\end{split}
\end{equation}
where $\odot$ denotes the Hadamard product.

Finally we  pass the visual-semantic features $\{{M}_{ab}\}$ to a fully-connected layer and a sigmoid layer to generate the visual-semantic score map $\{p_{ab}\}$. 
Each value $p_{a,b}$ in the visual-semantic score map denotes the predicted matching score of the temporal moment ($a,b$) for the sentence query.
The maximum of the score map $p$ corresponds to the grounding result for the sentence query.

The visual localizer directly guesses the most interested moments based on the visual feature of moment proposals $\{f^{V_{ab}}\}$ without the input of sentence query.
% aiming to learn video moment bias from the data. 
Specifically, the video moment features $\{f^{V_{ab}}\}$ are directly passed to a fully-connected layer and sigmoid layer to generate a  visual score map $\{p_{ab}^\prime\}$, which represents the predicted prior of video moment $(a,b)$ to be grounded.
 
\subsection{Sample Importance Reweighing}

Each training sample consists of an input video $V$, a sentence
query $S$ and the temporal annotation $T$ associated with the query. 
During training, we need to determine which temporal moment in the temporal-sentence score map corresponds to the annotations and train the model accordingly.  
Instead of hard label, we assign each moment proposal with a soft label according to its overlap with the annotations. 
Specifically, for each moment  in the temporal-sentence score map, we compute the IoU score $IoU_{ab}$ between its temporal boundary ($a,b$) and the annotation $T$.
Then a soft ground truth label $gt_{ab}$ is assigned to it according to $IoU_{ab}$:
    \begin{equation}
    gt_{ab} = 
    \begin{cases}
    0  & IoU_{ab}\leq \mu_{min},\\
	    \frac{IoU_{ab}-\mu_{min}}{\mu_{max}-\mu_{min}} & \mu_{min} < IoU_{ab} < \mu_{max},\\
	 1  & IoU_{ab}\geq \mu_{max},\\
	\end{cases}
	\label{eq:iou}
    \end{equation}
where $\mu_{min}$ and $\mu_{max}$ are two thresholds to customize the distribution of soft labels.

The visual localizer’s goal is to learn the video moment bias and predict the localization results only from visual modality.
For each training sample, the visual localizer can be trained with a binary cross entropy loss, which is defined as:
\begin{equation}\label{eq:viusal_loss}
\mathcal{L}_{v} = -\sum_{(a,b)\in\mathcal{C}}  \mathrm{gt}_{ab}\mathrm{log}(p_{ab}^\prime)
+(1-\mathrm{gt}_{ab})\mathrm{log}(1-p_{ab}^\prime),
\end{equation}
where $\mathcal{C} = \{(a,b)|1\leq a\leq b\leq N\}$ is the set of all valid moment proposal boundaries and $p_{ab}^\prime$ is the prediction output of visual localizer.

To train the visual-semantic localizer, previous works commonly train it with binary cross entropy loss similar to Eq.~\ref{eq:viusal_loss} as
\begin{equation}\label{eq:viusal_semantic_loss}
\mathcal{L}_{vs} = -\sum_{(a,b)\in\mathcal{C}} \mathrm{gt}_{ab}\mathrm{log}(p_{ab})
+(1-\mathrm{gt}_{ab})\mathrm{log}(1-p_{ab}),
\end{equation}
where $p_{ab}$ is the prediction of visual-semantic localizer.

Due to the superficial bias between video moments and ground-truth, the temporal grounding model trained with the  cross entropy loss tends to simply exploit the video modality to make a prediction, rather than jointly understand both video and language as claimed before.
To this end, we  use the prediction output $p_{ab}^\prime$ of the visual localizer  to reweigh the importance of training sample and adjust the loss function $\mathcal{L}_{vs}$ for the the visual-semantic localizer accordingly. Specifically, we first compute the cosine similarity $s$ of visual localizer prediction $p^\prime=\{p_{ab}^\prime\}$ and ground-truth $\mathrm{gt}=\{\mathrm{gt}_{ab}\}$ as
\begin{equation}\label{eq:viusal_semantic_loss_debias}
s = \frac{p^\prime\cdot \mathrm{gt}}{||p^\prime||_2   ||\mathrm{gt}||_2}
\end{equation}
Then  the adjusted loss $\mathcal{L}_{vs}^\prime$ for visual-semantic localizer is reweighted by $s$  as follows
\begin{equation}\label{eq:viusal_semantic_loss_debias}
\mathcal{L}_{vs}^\prime = (1-s^\alpha) \cdot \mathcal{L}_{vs},
\end{equation}
where $\alpha$ is a hyper-parameter to control the weight decay. Intuitively, high value of $s$ implies large probability of inferring the ground truth merely from the visual modal, implying tight relevance to the video moment biases. Following this intuition, Eq.~\ref{eq:viusal_semantic_loss_debias} suppresses their sample weight.

Last, we define the total loss function for Debias-TLL as:%denoted by
\begin{equation}\label{eq:tot_loss}
\mathcal{L}_t =  \mathcal{L}_{v} + \mathcal{L}_{vs}^\prime,
\end{equation}
which consists of the loss $\mathcal{L}_{v}$ for visual localizer and the adjusted loss $\mathcal{L}_{vs}^\prime$ for visual-semantic localizer.

With the final loss function $\mathcal{L}_t$, Debias-TLL can be trained in an end-to-end manner to mitigate video moment bias. At the inference stage, we drop the visual localizer and only use the visual-semantic localizer.

\section{Experiment}
\subsection{Dataset}
\noindent
\textbf{ActivityNet Captions}. 
It consists of 19,209 untrimmed videos with the annotation of sentence description and moment boundary. The contents of the videos are diverse. It is originally built for dense-captioning events~\cite{dense_cap} and lately introduced for temporal grounding.
It is the largest existing dataset in the field of temporal grounding. 
There are 37,417, 17,505, and 17,031 moment-sentence pairs in the training, validation and testing set, respectively.

\noindent
\textbf{Charades-STA}. It contains 9,848 videos of daily indoors activities. It is originally designed for action recognition and localization. Gao et al.~\cite{tall} extend the temporal annotation (\emph{i.e.}, labeling the start and end time of moments) of this dataset with language descriptions and name it as Charades-STA. There are 3,720 moment-sentence pairs in the testing set.

\noindent
\textbf{DiDeMo}. It was recently proposed
in~\cite{tempo}, specially for natural language moment retrieval in
open-world videos. 
DiDeMo contains 10,464 videos with 4,021 annotated moment query pairs in the testing set. 

\subsection{Evaluation Metrics}

The commonly-adopted evaluation metric in temporal grounding
is known to be ``Recall@$N$,IoU=$\theta$ ''. For each sentence query  we calculate the Intersection over Union (IoU) between a grounded temporal segment and the ground truth. ``Recall@$N$,IoU=$\theta$ '' represents the percentage of top $N$ grounded temporal segments that have at least one segment with higher IoU than $\theta$.
Following previous works~\cite{2dmap,scdm}, we report the results as $N \in \{ 1, 5\}$ with $\theta \in  \{0.5, 0.7 \}$ 
for ActivityNet Captions, Charades-STA and DiDeMo dataset. 

\subsection{Baseline Methods}
We compare our methods with several state-of-the-art methods listed as followings:
\textbf{CTRL}~\cite{tall}: Cross-model Temporal Regression Localizer.
\textbf{PFGA}~\cite{wacv}: Proposal-free Temporal Moment Localization using Guided Attention.
\textbf{SCDM}~\cite{scdm}: Semantic Conditioned Dynamic Modulation.
\textbf{2D-TAN}~\cite{2dmap}: 2D Temporal Adjacent Networks.
We further consider following 
methods: 
\textbf{random}: Randomly select the moment proposals.  
\textbf{TLL}: The model with identical archetecture to Debias-TLL, but trained by  commonly used binary cross entropy loss.

\subsection{Implementation Details}
we use pretrained CNN~\cite{i3d} as previous methods to extract I3D video features on all datasets
And we use Glove~\cite{glove} word embeddings pretrained on Common Crawl to represent each word in the sentences. 
A three layer LSTM is applied to word-embeddings to obtain the sentence representation. 
The channel numbers of sentence feature and video proposal feature $d^S,d^V$ are all set to $512$ .The number of sampled clips $N$ is set to $32$.
For BM operations in the video encoder, we set sampling number  of each proposals to $32$.

During training, We use Adam~\cite{adam} with learning rate of $1\times10^{-4}$, a momentum of $0.9$ and batch size of $4$. $\alpha=1.0$ in Eq.~\ref{eq:viusal_semantic_loss_debias}.
During inference, we choose the moment proposals with the highest confidence score for the sentence query as the final result. 
If it is desired to select multiple moment locations per sentence (\emph{i.e.}, for R@5), non-maximum suppression (NMS) with a threshold of $0.4$ is applied to remove redundant candidates.

\subsection{Analysis of Video Moment Biases}\label{sec:unimodal}
To show the video moment bias in the temporal grounding model,
in this experiment, we mask all the words of the sentence input and evaluate existing models with single video input (marked as ``video-only") on the ActivityNet Captions. 
The results are summarized in Table~\ref{table:unimodal_acnet_res}, where all of these methods achieve better performance than the random method with large margins. 
This shows that the model can heavily exploit the superficial correlation between video moment and ground-truth to provide correct localization results.

%%%%%%%%%%%%%%%% Unimodal Results on ActivityNet Caption %%%%%%%%%%%%%%%%
\begin{table}[t]
\centering
\caption{Performance evaluation results on the ActivityNetCap.
}\label{table:unimodal_acnet_res}
\scalebox{0.78}{
\begin{tabular}{c|c|ccccccc}
\hline
\hline
 \multirow{2}{*}{Input} &\multirow{2}{*}{Method}  &R@1 &R@1 &R@5 &R@5\\
 &&IoU=0.5 &IoU=0.7 &IoU=0.5 &IoU=0.7\\
\hline
 &random &13.99 &4.69  &44.69 &17.64\\ 
\hline 
\multirow{5}{*}{video \& query}
&CTRL		&29.01  &10.34 	&59.17 &37.54\\
&PFGA 		&33.04  &19.26  &- &-\\
&SCDM    	&36.75  &19.86  &64.99 &41.53\\
&2D-TAN  	&44.51  &26.54  &77.13 &61.96\\
&TLL     	&44.24 &27.01  &75.22 &60.23\\
\hline 
\multirow{4}{*}{video-only} 
&PFGA 		&21.69 	&12.56  &- &-\\
&SCDM    	&23.84  &12.93  &51.66 &32.36\\
&2D-TAN  	&27.56 	&13.93  &61.65 &36.78\\
&TLL     	&28.10 &13.96 &59.07 &36.25\\
\hline 
\end{tabular}
}
\end{table}

%Cross Dataset Performance
\begin{table}[t]
\centering
\caption{Cross-scenario performance of video-only model on the AcNet2Charades and AcNet2DiDeMo.  
}\label{table:cross_performance_base}
\scalebox{0.8}{
\begin{tabular}{c|c|ccccccc}
\hline
\hline
 \multirow{2}{*}{Dataset} &\multirow{2}{*}{Method}  &R@1 &R@1 &R@5 &R@5\\
 &&IoU=0.5 &IoU=0.7 &IoU=0.5 &IoU=0.7\\
\hline 
\multirow{2}{*}{Charades}
&random  &11.88 &3.76 &46.64 &16.88 \\
&video-only    &6.68 &0.41 &56.68 &25.55\\
\hline 
\multirow{2}{*}{DiDemo}
&random & 8.36 & 2.59 &37.53 &11.79\\
&video-only   &4.87 &1.22 &38.74 &16.54\\
\hline 
\end{tabular}
}
\end{table}

%%%%%%%%%%%%%%%% Unimodal Bias Analysis %%%%%%%%%%%%%%%%
\begin{figure}[t!] 
\begin{subfigure}{0.23\textwidth}
\centering
\includegraphics[width=0.95\linewidth]{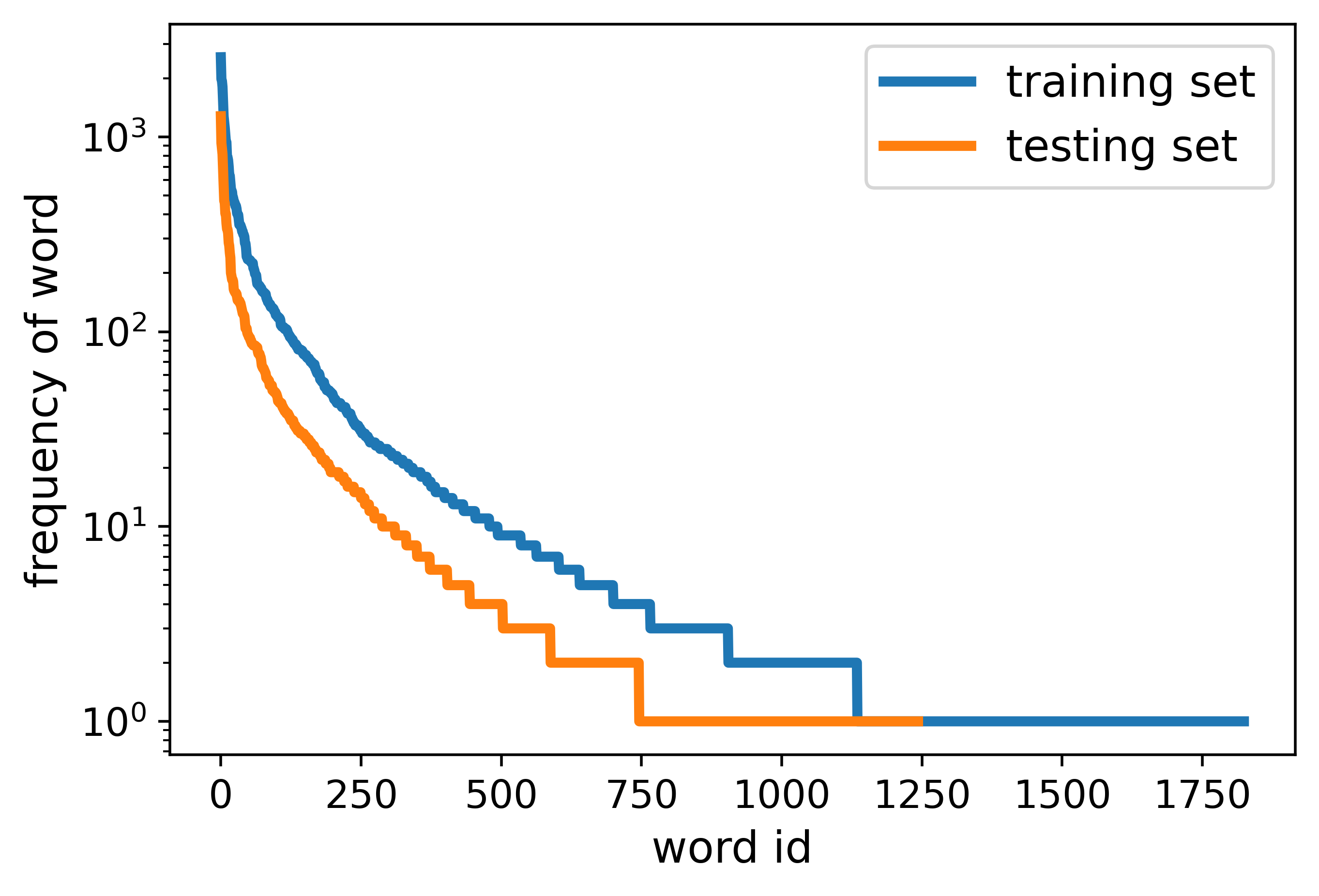}
\vspace{-0.25cm}
\caption{Frequency of actions.}\label{fig:action_freq1}
\end{subfigure}
\begin{subfigure}{0.23\textwidth}
\centering
\includegraphics[width=0.95\linewidth]{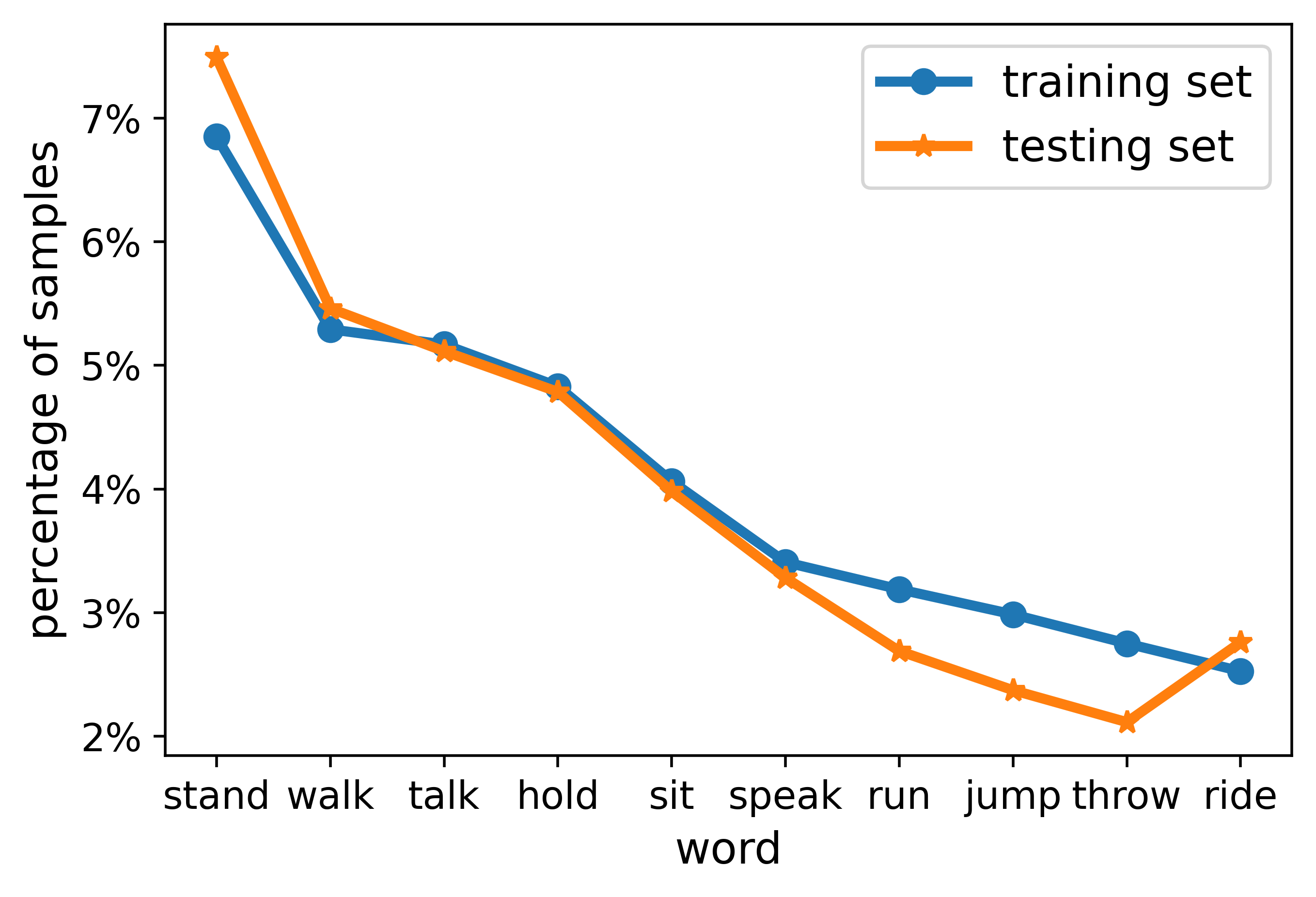} 
\vspace{-0.25cm}
\caption{Top frequency of actions.}\label{fig:action_freq2}
\end{subfigure}

\begin{subfigure}{0.23\textwidth}
\centering
\includegraphics[width=0.95\linewidth]{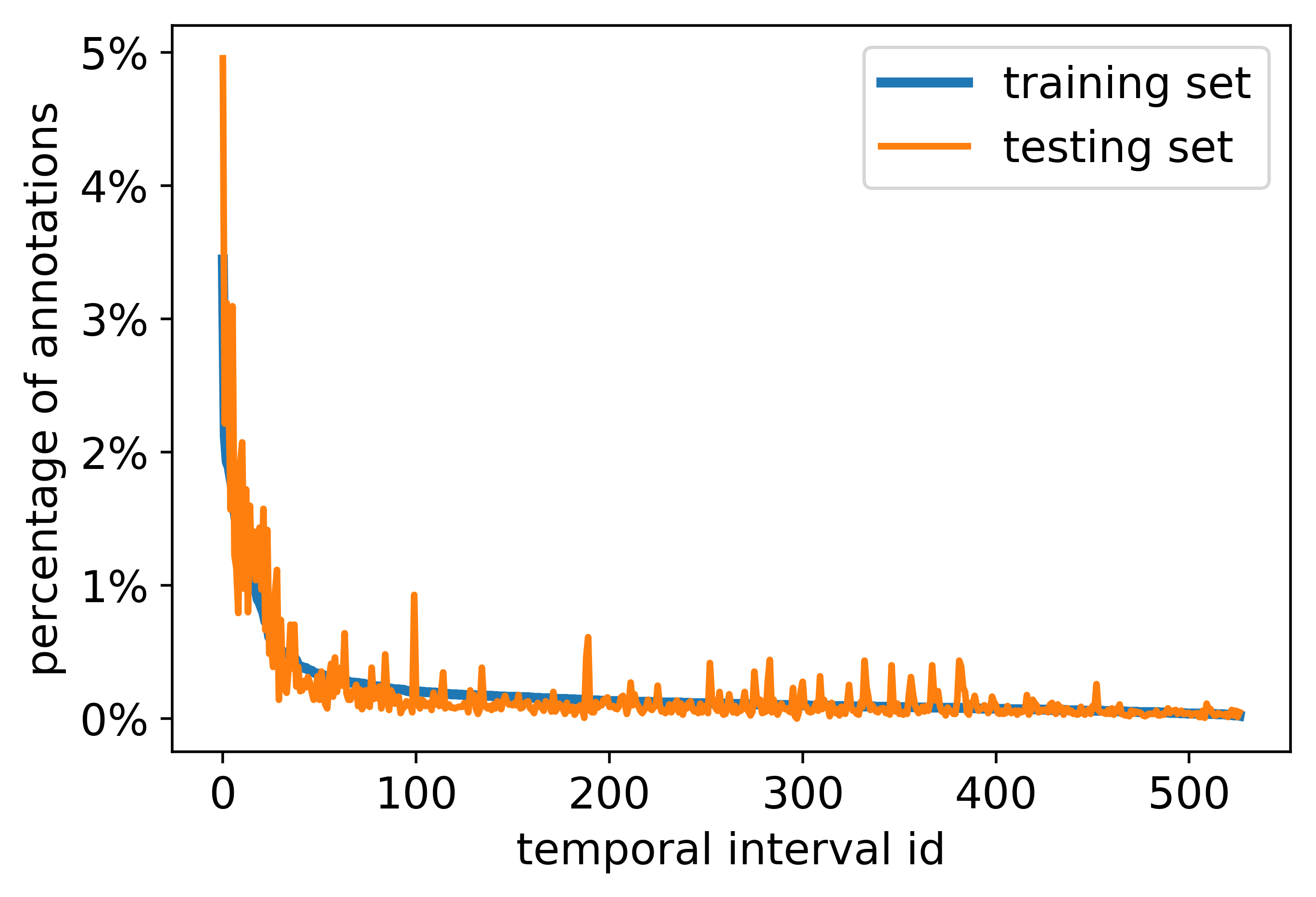}
\vspace{-0.25cm}
\caption{Frequency of intervals.}\label{fig:proposal_freq1}
\end{subfigure}
\begin{subfigure}{0.23\textwidth}
\centering
\includegraphics[width=0.95\linewidth]{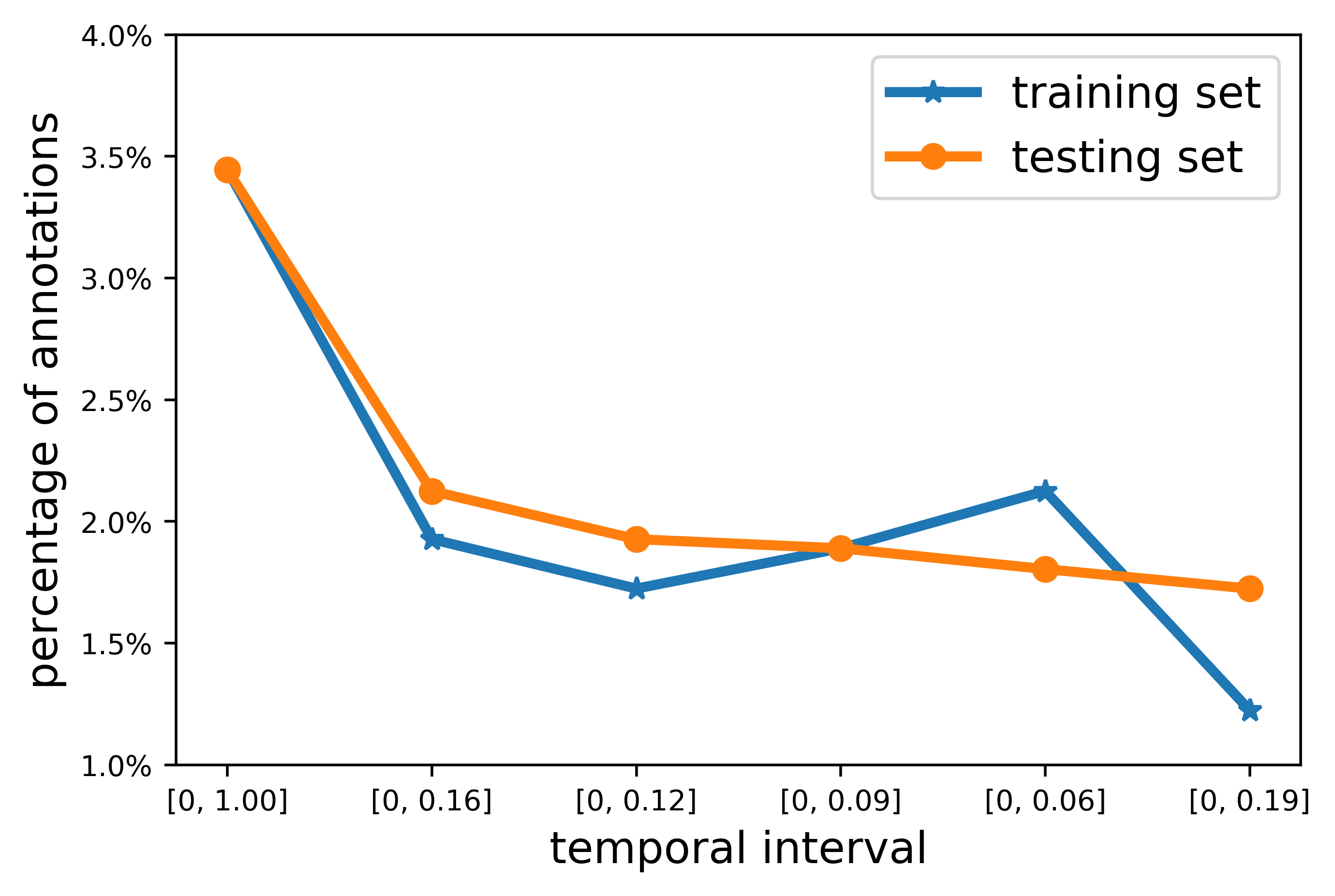} 
\vspace{-0.25cm}
\caption{Top frequency of intervals.}\label{fig:proposal_freq2}
\end{subfigure}
\vspace{-0.2cm}
\caption{Video moment bias analysis on ActivityNet Captions}\label{fig:unimodal}
\vspace{-0.4cm}
\end{figure}

Here we further identify two sorts of specific biases (visual content bias and temporal interval bias) that the model can exploit to localize the target moment when ignoring the sentence query. Figure~\ref{fig:action_freq1} presents the frequencies of action concepts in all the sentence queries in both training and testing data of the ActivityNet Captions. The action concept distribution follows a long-tail distribution, and some actions concept are much more frequently queried than others. We further illustrate the top frequency of action concepts in Fig.~\ref{fig:action_freq2}, showing that training and testing data share a similar distribution. A similar conclusion also exists in the object concepts in the sentence query.
Furthermore, we illustrate the frequency distribution and the top frequency of temporal intervals of the queried moments in Figure~\ref{fig:proposal_freq1} and~\ref{fig:proposal_freq2}. Like the video content, the temporal intervals also follow long-tail distributions shared by the training and testing data. The temporal information of the moment can be captured by the modern temporal grounding model with temporal context modeling using recurrent neural networks, non-local blocks, etc.
This means video moment proposals that contain certain video concepts and temporal intervals are more likely to be localized as positive, and the model can then infer the localization results only according to the video, irrespective of the sentence query.

\subsection{Cross-Scenario Evaluation}

To quantify the effect of all afore-mentioned task-specific biases, we evaluate under a cross-scenario setting for temporal grounding in video. The training and evaluation processes are conducted on two different distributions where video moment biases cannot be transferred from one data distribution to another. 
Specifically, we train the model on the ActivityNet Captions considering its large-scale data amount and the diversity of scenes and activities.  Then we test the model on the Charades-STA and DiDeMo (dubbed as AcNet2Charades and AcNet2DiDeMo repectively). 
As shown in Table~\ref{table:cross_performance_base}, under the cross-scenario setting, 
the ``video-only'' TLL model fails to perform well on the testing data and is even inferior than the random guess model on the metrics of R1.

%%%%%%%%%%%%%%%%%%%%%%%%%%%%% Results on AcNet2Charades %%%%%%%%%%%%%%%%%%%%%%%%%%%%% 
\begin{table}[t]
\centering
\caption{Performance evaluation results on the AcNet2Charades. 
}\label{table:AcNet2Charades_res}
\scalebox{0.9}{
\begin{tabular}{c|ccccccc}
\hline
\hline
 \multirow{2}{*}{Method}&R@1 &R@1 &R@5 &R@5  \\
 &   IoU=0.5 &  IoU=0.7   &IoU=0.5 &  IoU=0.7 \\
\hline
random  &11.88 &3.76 &46.64 &16.88 \\
video-only     	&6.68 &0.41 &56.68 &25.55\\
PFGA  &5.75 &1.53 &- &- \\
SCDM  &15.91 &6.19 &54.04 &30.39 \\
2D-TAN  &15.81 &6.30 &59.06 &31.53 \\
Debias-TLL &\textbf{21.45} &\textbf{10.38} &\textbf{62.34} &\textbf{32.90}\\
\hline 
\end{tabular}
}
\end{table}

%%%%%%%%%%%%%%%%%%%%%%%%%%%%% Results on AcNet2DiDeMo %%%%%%%%%%%%%%%%%%%%%%%%%%%%% 
\begin{table}[t]
\centering
\caption{
Performance evaluation results on the AcNet2DiDeMo.
}
\label{table:AcNet2DiDeMo_res}
\scalebox{0.9}{
\begin{tabular}{c|ccccc}
\hline
\hline
 \multirow{2}{*}{Method} &R@1 &R@1 &R@5  &R@5\\
 &    IoU=0.5 &  IoU=0.7 & IoU=0.5 &  IoU=0.7\\
\hline
random & 8.36 & 2.59 &37.53 &11.79\\
video-only     	&4.87 &1.22 &38.74 &16.54\\
PFGA &6.24 &2.01 &- &-\\
SCDM &10.88 &4.34 &43.30 &18.40\\ 
2D-TAN &12.50  &5.50  &44.88 &20.73\\
Debias-TLL &\textbf{13.11} &\textbf{7.70} &\textbf{44.98} &\textbf{21.32}\\
\hline 
\end{tabular}
}
\end{table}

\subsection{Performance Comparison}
The results of the proposed Debias-TLL and the baselines on ACNet2Charades and ACNet2DiDeMo are summarized in Table~\ref{table:AcNet2Charades_res} and~\ref{table:AcNet2DiDeMo_res} respectively.
Our algorithm outperforms all the competing methods with a clear margin. It is noticeable
that the proposed technique surpasses the state-of-the-art performances by 8.64\% and 4.08\% points in terms of R1@0.5 and R1@0.7 metric, respectively. This verifies the effectiveness of the
sample importance reweighing in cross-scenario temporal grounding. The prevailing solutions for temporal grounding can be grouped into two categories i.e. top-down and bottom-up approach. We note that the top-down method PFGA achieves much inferior results than the top-down methods SCDM and 2D-TAN, which suggests the superiority of top-down design compared to the bottom-up one under the cross-scenario setting. We suspect that this is because the bottom-up approach directly predicts each frame's probabilities as ground-truth interval boundary and more easy to overfit to the temporal intervals bias.

\subsection{Ablation Study}
\noindent
\textbf{Impact of Importance Reweighing} \indent
To study the impact of sample importance reweighing in Debias-TLL, we substitute the two-model based adjusted loss Eq.~\ref{eq:viusal_semantic_loss_debias} with the commonly used binary cross entropy loss (marked as TLL), accordingly train the model TLL on ActivityNet Captions.
The evaluation results on both Chrades-STA and DiDeMo are listed in Table~\ref{table:abl_of_debias}. As expected, without the sample importance reweighing, the TLL model gets inferior results than Debias-TLL with a clear margin on both datasets, verifying the effectiveness of the proposed technique under the setting of cross-scenario temporal grounding.

\begin{figure}[t]
\includegraphics[width=0.23\textwidth]{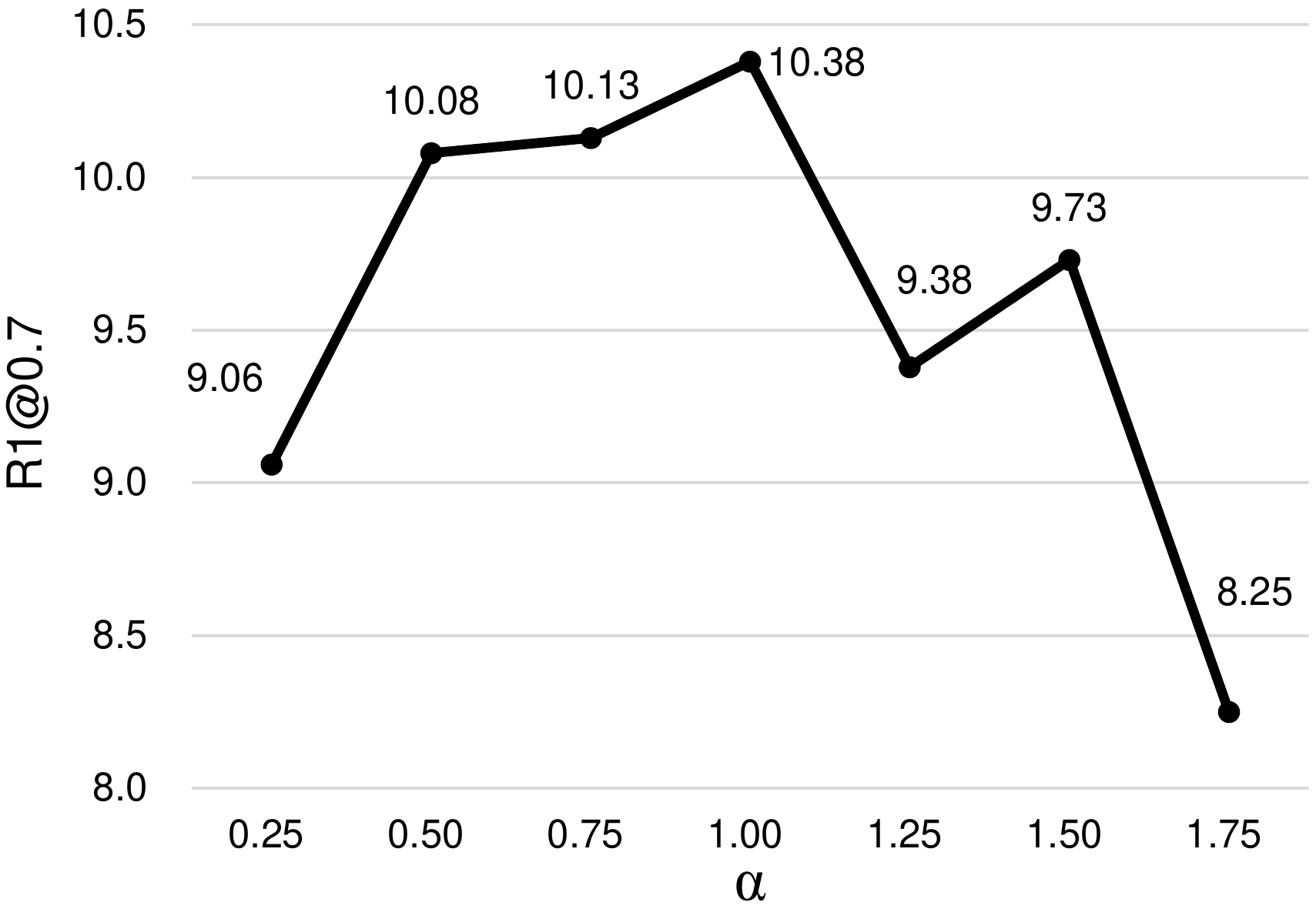}
\includegraphics[width=0.23\textwidth]{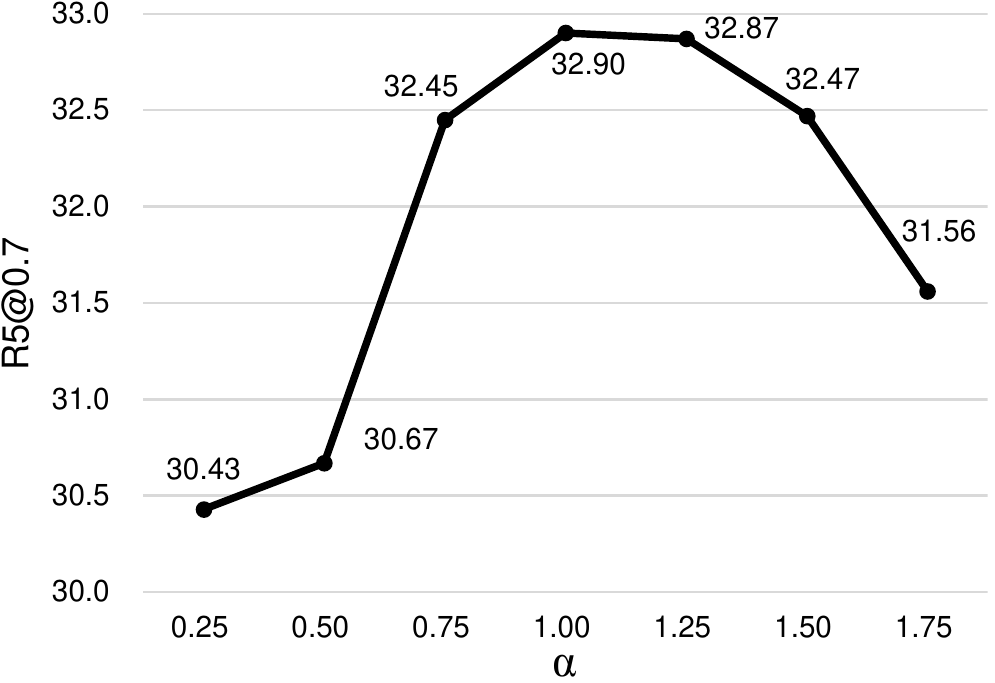}
\caption{Impact of $\alpha$ on the AcNet2Charades.}\label{fig:abl_of_alpha}
\vspace{-0.1in}
\end{figure}

\noindent
\textbf{Impact of Hyperparameter $\alpha$} \indent
The hyperparameter $\alpha$ plays a key role in controlling the magnitude of sample importance reweighing.   
And when $\alpha\rightarrow\infty$, $1-s^\alpha$ approximates $1$ and then the adjusted loss Eq.~\ref{eq:viusal_semantic_loss_debias}  approximates to binary cross entropy loss Eq.~\ref{eq:viusal_semantic_loss}. An appropriate setting of $\alpha$ is required to rebalance the heavily video moment biased data.
Here we study the effect of different settings of $\alpha$. 
Figure~\ref{fig:abl_of_alpha} illustrates the performance R1@0.7 and R5@0.7 of the Debias-TLL model on AcNet2Charades with $\alpha$ set to 0.25 to 1.0. We found that the performances increase until $\alpha=1.0$ and then decrease afterward. This shows that $\alpha=1$ is a proper selection to balance the distribution of training samples and achieve satisfactory results. Note that even when $\alpha=1.75$, the performance is still much superior to the TTL baseline without sample importance reweighing. 

%Ablation Study of Debias
\begin{table}[t]
\centering
\caption{Ablation study on AcNet2Charades and AcNet2DiDeMo. 
}\label{table:abl_of_debias}
\scalebox{0.8}{
\begin{tabular}{c|c|ccccccc}
\hline
\hline
 \multirow{2}{*}{Dataset} &\multirow{2}{*}{Method}  &R@1 &R@1 &R@5 &R@5\\
 &&IoU=0.5 &IoU=0.7 &IoU=0.5 &IoU=0.7\\
\hline 
\multirow{2}{*}{Charades}
&TLL &14.76 &6.12 &60.41 &31.89\\
&Debias-TLL &\textbf{21.45} &\textbf{10.38} &\textbf{62.34} &\textbf{32.90}\\
\hline 
\multirow{2}{*}{DiDeMo}
&TLL &10.25 &5.13 &44.73 &19.49\\
&Debias-TLL &\textbf{13.11} &\textbf{7.70} &\textbf{44.98} &\textbf{21.32}\\
\hline 
\end{tabular}
}
\end{table}

%%%%%%%%%%%%%%%% Ablation Figure of Alpha %%%%%%%%%%%%%%%%

\section{Conclusion}
In this paper, we show that temporal grounding models are heavily affected by video moment bias of the data, limiting the generalization performance on cross-scenario testing data. 
To prevent the model from naively memorizing the biases and enforce it to ground the query sentence based on true cross-modal understanding, we propose a novel Debiased Temporal Language Localizer with a two-model based data-reweighing mechanism.
Experiments show large-margin superiority of the proposed method in comparison with state-of-the-art competitors in cross-scenario temporal grounding.

\bibliographystyle{named}
\bibliography{ijcai21.bib}

\begin{thebibliography}{}

\bibitem[\protect\citeauthoryear{Anne~Hendricks \bgroup \em et al.\egroup
  }{2017}]{mcn}
Lisa Anne~Hendricks, Oliver Wang, Eli Shechtman, Josef Sivic, Trevor Darrell,
  and Bryan Russell.
\newblock Localizing moments in video with natural language.
\newblock In {\em ICCV}, 2017.

\bibitem[\protect\citeauthoryear{Cadene \bgroup \em et al.\egroup
  }{2019}]{vqa_unimodal}
Remi Cadene, Corentin Dancette, Matthieu Cord, Devi Parikh, et~al.
\newblock Rubi: Reducing unimodal biases for visual question answering.
\newblock In {\em NeurIPS}, 2019.

\bibitem[\protect\citeauthoryear{Carreira and Zisserman}{2017}]{i3d}
Joao Carreira and Andrew Zisserman.
\newblock Quo vadis, action recognition? a new model and the kinetics dataset.
\newblock In {\em CVPR}, 2017.

\bibitem[\protect\citeauthoryear{Chen \bgroup \em et al.\egroup
  }{2020}]{vqa_counterfact}
Long Chen, Xin Yan, Jun Xiao, Hanwang Zhang, Shiliang Pu, and Yueting Zhuang.
\newblock Counterfactual samples synthesizing for robust visual question
  answering.
\newblock 2020.

\bibitem[\protect\citeauthoryear{Gao \bgroup \em et al.\egroup }{2017}]{tall}
Jiyang Gao, Chen Sun, Zhenheng Yang, and Ram Nevatia.
\newblock Tall: Temporal activity localization via language query.
\newblock In {\em ICCV}, 2017.

\bibitem[\protect\citeauthoryear{Ghosh \bgroup \em et al.\egroup }{2019}]{excl}
Soham Ghosh, Anuva Agarwal, Zarana Parekh, and Alexander Hauptmann.
\newblock Excl: Extractive clip localization using natural language
  descriptions.
\newblock {\em arXiv preprint arXiv:1904.02755}, 2019.

\bibitem[\protect\citeauthoryear{Hendricks \bgroup \em et al.\egroup
  }{2018}]{tempo}
Lisa~Anne Hendricks, Oliver Wang, Eli Shechtman, Josef Sivic, Trevor Darrell,
  and Bryan Russell.
\newblock Localizing moments in video with temporal language.
\newblock In {\em EMNLP}, 2018.

\bibitem[\protect\citeauthoryear{Hochreiter and Schmidhuber}{1997}]{lstm}
Sepp Hochreiter and J{\"u}rgen Schmidhuber.
\newblock Long short-term memory.
\newblock {\em Neural computation}, 1997.

\bibitem[\protect\citeauthoryear{Jeffrey~Pennington and Manning}{2014}]{glove}
RichardSocher Jeffrey~Pennington and ChristopherD Manning.
\newblock Glove: Global vectors for word representation.
\newblock In {\em EMNLP}, 2014.

\bibitem[\protect\citeauthoryear{Kingma and Ba}{2014}]{adam}
Diederik~P Kingma and Jimmy Ba.
\newblock Adam: A method for stochastic optimization.
\newblock {\em arXiv preprint arXiv:1412.6980}, 2014.

\bibitem[\protect\citeauthoryear{Krishna \bgroup \em et al.\egroup
  }{2017}]{dense_cap}
Ranjay Krishna, Kenji Hata, Frederic Ren, Li~Fei-Fei, and Juan Carlos~Niebles.
\newblock Dense-captioning events in videos.
\newblock In {\em ICCV}, 2017.

\bibitem[\protect\citeauthoryear{Lin \bgroup \em et al.\egroup }{2019}]{bmn}
Tianwei Lin, Xiao Liu, Xin Li, Errui Ding, and Shilei Wen.
\newblock Bmn: Boundary-matching network for temporal action proposal
  generation.
\newblock In {\em ICCV}, 2019.

\bibitem[\protect\citeauthoryear{Liu \bgroup \em et al.\egroup
  }{2018a}]{lifeifei}
Bingbin Liu, Serena Yeung, Edward Chou, De-An Huang, Li~Fei-Fei, and Juan
  Carlos~Niebles.
\newblock Temporal modular networks for retrieving complex compositional
  activities in videos.
\newblock In {\em ECCV}, 2018.

\bibitem[\protect\citeauthoryear{Liu \bgroup \em et al.\egroup }{2018b}]{acrn}
Meng Liu, Xiang Wang, Liqiang Nie, Xiangnan He, Baoquan Chen, and Tat-Seng
  Chua.
\newblock Attentive moment retrieval in videos.
\newblock In {\em SIGIR}, 2018.

\bibitem[\protect\citeauthoryear{Liu \bgroup \em et al.\egroup
  }{2018c}]{attention_2}
Meng Liu, Xiang Wang, Liqiang Nie, Qi~Tian, Baoquan Chen, and Tat-Seng Chua.
\newblock Cross-modal moment localization in videos.
\newblock In {\em ACM MM}, 2018.

\bibitem[\protect\citeauthoryear{Qi \bgroup \em et al.\egroup
  }{2020}]{visual_dialog}
Jiaxin Qi, Yulei Niu, Jianqiang Huang, and Hanwang Zhang.
\newblock Two causal principles for improving visual dialog.
\newblock In {\em CVPR}, 2020.

\bibitem[\protect\citeauthoryear{Ramakrishnan \bgroup \em et al.\egroup
  }{2018}]{vqa_regularization}
Sainandan Ramakrishnan, Aishwarya Agrawal, and Stefan Lee.
\newblock Overcoming language priors in visual question answering with
  adversarial regularization.
\newblock In {\em NeurIPS}, 2018.

\bibitem[\protect\citeauthoryear{Rodriguez \bgroup \em et al.\egroup
  }{2020}]{wacv}
Cristian Rodriguez, Edison Marrese-Taylor, Fatemeh~Sadat Saleh, Hongdong Li,
  and Stephen Gould.
\newblock Proposal-free temporal moment localization of a natural-language
  query in video using guided attention.
\newblock In {\em WACV}, 2020.

\bibitem[\protect\citeauthoryear{Stroud \bgroup \em et al.\egroup
  }{2019}]{comp}
Jonathan~C Stroud, Ryan McCaffrey, Rada Mihalcea, Jia Deng, and Olga
  Russakovsky.
\newblock Compositional temporal visual grounding of natural language event
  descriptions.
\newblock {\em arXiv preprint arXiv:1912.02256}, 2019.

\bibitem[\protect\citeauthoryear{Tang \bgroup \em et al.\egroup
  }{2020}]{unbias_scene_graph}
Kaihua Tang, Yulei Niu, Jianqiang Huang, Jiaxin Shi, and Hanwang Zhang.
\newblock Unbiased scene graph generation from biased training.
\newblock In {\em CVPR}, 2020.

\bibitem[\protect\citeauthoryear{Wang \bgroup \em et al.\egroup }{2019a}]{rl}
Weining Wang, Yan Huang, and Liang Wang.
\newblock Language-driven temporal activity localization: A semantic matching
  reinforcement learning model.
\newblock In {\em CVPR}, 2019.

\bibitem[\protect\citeauthoryear{Wang \bgroup \em et al.\egroup
  }{2019b}]{sm_rl}
Weining Wang, Yan Huang, and Liang Wang.
\newblock Language-driven temporal activity localization: A semantic matching
  reinforcement learning model.
\newblock In {\em CVPR}, 2019.

\bibitem[\protect\citeauthoryear{Wang \bgroup \em et al.\egroup }{2020}]{cbp}
Jingwen Wang, Lin Ma, and Wenhao Jiang.
\newblock Temporally grounding language queries in videos by contextual
  boundary-aware prediction.
\newblock In {\em AAAI}, 2020.

\bibitem[\protect\citeauthoryear{Yuan \bgroup \em et al.\egroup }{2019}]{scdm}
Yitian Yuan, Lin Ma, Jingwen Wang, Wei Liu, and Wenwu Zhu.
\newblock Semantic conditioned dynamic modulation for temporal sentence
  grounding in videos.
\newblock In {\em NeurIPS}, 2019.

\bibitem[\protect\citeauthoryear{Zhang \bgroup \em et al.\egroup }{2019a}]{man}
Da~Zhang, Xiyang Dai, Xin Wang, Yuan-Fang Wang, and Larry~S Davis.
\newblock Man: Moment alignment network for natural language moment retrieval
  via iterative graph adjustment.
\newblock In {\em CVPR}, 2019.

\bibitem[\protect\citeauthoryear{Zhang \bgroup \em et al.\egroup
  }{2019b}]{mm20}
Songyang Zhang, Jinsong Su, and Jiebo Luo.
\newblock Exploiting temporal relationships in video moment localization with
  natural language.
\newblock In {\em ACM MM}, 2019.

\bibitem[\protect\citeauthoryear{Zhang \bgroup \em et al.\egroup
  }{2019c}]{cmin}
Zhu Zhang, Zhijie Lin, Zhou Zhao, and Zhenxin Xiao.
\newblock Cross-modal interaction networks for query-based moment retrieval in
  videos.
\newblock In {\em SIGIR}, 2019.

\bibitem[\protect\citeauthoryear{Zhang \bgroup \em et al.\egroup
  }{2020}]{2dmap}
Songyang Zhang, Houwen Peng, Jianlong Fu, and Jiebo Luo.
\newblock Learning 2d temporal adjacent networks for moment localization with
  natural language.
\newblock In {\em AAAI}, 2020.

\end{thebibliography}

\end{document}